\documentclass{article}

\usepackage[dblblindworkshop, final]{neurips_2026}
\workshoptitle{Real-Time Multimodal Conversational AI (ReMuCAI)}

\usepackage[utf8]{inputenc}
\usepackage[T1]{fontenc}
\usepackage{hyperref}
\usepackage{url}
\usepackage{booktabs}        
\usepackage{amsfonts,amsmath}
\usepackage{graphicx}
\usepackage{multirow}
\usepackage{xcolor}       
\usepackage{pgfplots}     
\pgfplotsset{compat=1.18} 
\usepackage{amssymb}      

\usepackage{listings}     
\lstdefinestyle{promptstyle}{
  basicstyle=\ttfamily\scriptsize,
  breaklines=true,
  breakatwhitespace=false,
  columns=fullflexible,
  keepspaces=true,
  frame=single,
  framesep=3pt,
  xleftmargin=3pt,
  xrightmargin=3pt,
  backgroundcolor=\color{gray!5},
  captionpos=b,
  showstringspaces=false
}
\title{An Agentic Approach with Evidence-Grounded Verification for Multimodal Emotion Reasoning in Conversation}

\author{
Elyas Rashno\thanks{Corresponding author: \texttt{elyas.rashno@queensu.ca}} \\
School of Computing, \\ Queen's University, \\
Kingston, Ontario, Canada
\And
Farhana Zulkernine \\
School of Computing, \\ Queen's University, \\
Kingston, Ontario, Canada
}

\begin{document}

\maketitle


\begin{abstract}
LLM-based multimodal emotion reasoning systems perform well on standard
benchmarks, but improving them through instruction tuning can be
computationally expensive. Agentic pipelines built from frozen models avoid
this cost, but introduce a different challenge: when a prediction is wrong,
the system must identify which agent or stage caused the error. We address
this problem with an agentic framework for traceable multimodal emotion
reasoning and prompt-space adaptation. Modality-specific perception agents
analyze visual, acoustic, and lexical inputs and post evidence-cited
observations to a shared board. An integration agent combines these
observations with retrieved multimodal examples to produce an emotion
description, label set, and supporting citations, while an MLLM critic can
request one revision before the output is finalized. For incorrect
predictions, a separate evidence-grounded verification stage applies
deterministic checks to verify that cited evidence and references are valid,
and an entailment check to determine whether the final claims are supported
by that evidence. Detected failures are then grouped by the responsible
component and failure type, converted into targeted prompt patches, and
accepted only after validation on a held-out set. This adapts the agent
prompts without updating model weights. On the MER2025 dataset, targeted
prompt adaptation outperforms the frozen static pipeline on all four
MLLMs, ranging from 2B to 38B parameters, and matches or outperforms
undifferentiated reflection baselines on each of them: VideoLLaMA3-2B, Phi-3.5-vision-instruct,
Qwen2.5-VL-7B-Instruct, and InternVL2.5-38B. It improves F-score by 0.6--2.2 points depending on the
backbone, including an increase from 52.3 to 54.5 on Qwen2.5-VL-7B.
\end{abstract}




\section{Introduction}
\label{sec:intro}

Multimodal emotion reasoning asks a system to infer a speaker's emotional state from three signals at once \citep{poria2017review}, including what they say, how they say it, and how they look while saying it. The task began as a single-channel exercise, with early systems reading facial expression \citep{lideng2022facial}, vocal prosody \citep{schuller2018speech}, or text \citep{devlin2019bert} in isolation, on the premise that emotion leaves a legible trace in each modality on its own. But single channels proved unreliable because a smile can mask distress, while a flat voice can carry it. The field therefore turned to fusion, combining modalities through learned feature-level alignment \citep{he2022dynamic, tsai2020multimodal}. Fusion improved accuracy and broadened the task's reach into human-computer interaction \citep{cowie2001emotion}, affective-state monitoring \citep{picard2001toward}, and sentiment analysis of online video and social media \citep{soleymani2017survey}. Recent work extends the same machinery toward unified emotional understanding and generation \citep{zhu2026uniemo}. Across all three stages, however, the task has been treated the same way by mapping features to a label. Yet emotion perception is not a lookup. It is an inference problem, and an inference problem grows harder as the signal grows richer.

Multimodal large language models (MLLMs) changed this picture by replacing learned feature-level mappings with generative models that can reason about affect in free text rather than emit a fixed label \citep{cheng2024emotionllama}. Instruction-tuned systems learn this capability through model weights and are evaluated mainly on their final labels or descriptions \citep{lian2025affectgpt}. More recent work instead composes frozen MLLMs into multi-agent pipelines, using multiple agents to perceive, retrieve, reflect, and verify affective evidence \citep{yang2025mermaid, wang2026affectagent}. These approaches move emotion recognition toward explicit reasoning, but they still leave a fundamental verification problem. When an agent makes an incorrect affective claim, its correctness is typically judged by another model rather than checked against the visual, acoustic, or lexical signal that supports the claim. The system can therefore produce a plausible explanation without establishing whether its evidence is real or whether the conclusion follows from that evidence.

We address this gap by making affective claims explicitly traceable and independently re-checkable. Three modality-specialist agents record visual, acoustic, and lexical observations with citations to a specific frame, time span, or transcript span, and an integration agent combines these with retrieved precedent into a draft description, aided by an inline MLLM critic that can request one revision before the draft is finalized. A separate verification layer then audits the finalized draft using deterministic checks that re-compute claims against the cited raw signals and an entailment check that tests whether the description follows from its evidence, attributing any failure to the stage that caused it rather than to another model's judgment. A verifier-guided repair loop clusters recurring failures by agent and defect type, synthesizes targeted prompt patches, and admits a patch only after validation on held-out data. On MER2025 Track 3 \citep{lian2025mer2025}, targeted repair outperforms a static pipeline on all four MLLMs from 2B to 38B parameters and matches or outperforms two undifferentiated reflection baselines on each, and we further examine the mechanism through component ablations and a data-efficiency study. 

The remainder of the paper is organized as follows. Section~\ref{sec:related} reviews prior work on MLLM approaches for multimodal emotion reasoning. Section~\ref{sec:method} presents the proposed method. Section~\ref{sec:results} reports the experimental results and ablation studies. Section~\ref{sec:limitations} states the limitations. Section~\ref{sec:conclusion} concludes the paper and outlines future work.

\section{Related Work}
\label{sec:related}

\subsection{LLMs and MLLMs for Emotion Understanding}

\paragraph{Parameter-frozen prompting.}
Recent surveys organize LLM- and MLLM-based emotion methods into
\emph{parameter-frozen} and \emph{parameter-tuning} paradigms
\citep{shou2025mllmsurvey}. In the parameter-frozen setting, the backbone
weights remain unchanged and the task is specified through zero-shot
instructions or in-context examples. \citet{feng2024arcllm}
study text-based LLMs for affect recognition in conversations using IEMOCAP,
EmoWOZ, and DAIC-WOZ. They evaluate zero-shot prompting and few-shot
in-context learning, and also examine robustness to ASR-corrupted transcripts.
\citet{etesam2024contextual} study contextual emotion recognition
from images on EMOTIC using two approaches: image captioning followed by a
language-only LLM, and direct prediction with a large vision-language model.
They evaluate both zero-shot and fine-tuned settings. These studies show that
general-purpose language and vision-language models can be applied to affect
tasks through prompting, but their predictions are produced directly from the
provided context rather than from an explicit modality-specific evidence trace
that can later be independently audited.

\paragraph{Parameter-tuning approaches.}
The parameter-tuning paradigm adapts model parameters using task-specific or
emotion-oriented supervision. \citet{feng2024arcllm} additionally
fine-tune text-based LLMs for emotion recognition in conversation, while
\citet{etesam2024contextual} fine-tune a vision-language model for
contextual emotion recognition on EMOTIC. Emotion-LLaMA
\citep{cheng2024emotionllama} extends this direction to multimodal emotion
recognition and reasoning by integrating audio, visual, and textual inputs with
emotion-specific encoders, aligning their features in a shared space, and using
instruction tuning with a modified LLaMA backbone. AffectGPT
\citep{lian2025affectgpt} introduces the MER-Caption dataset together with a
multimodal model that uses pre-fusion operations before language-model
reasoning, and evaluates the resulting system through the MER-UniBench
benchmark. These methods improve affect recognition and reasoning through
learned specialization and multimodal integration, but they do not provide a
mechanism that independently re-derives each affective claim from its cited raw
signal and attributes a detected failure to a specific processing component.

\paragraph{Multi-agent emotion reasoning.}
A more recent line decomposes emotion reasoning across multiple agents.
MERMAID \citep{yang2025mermaid} is an image-based multi-agent framework with
a decision agent, textual and visual self-reflection agents, and an
emotion-guided augmentation agent. Its cross-modal verification repeatedly
checks a prediction from one view using reflection from another view and feeds
the resulting feedback back to the decision agent. AffectAgent
\citep{wang2026affectagent} instead targets retrieval-augmented multimodal
emotion recognition with three specialized agents: a query planner, an
evidence filter, and an emotion generator. These agents are jointly optimized
with multi-agent proximal policy optimization, while modality balancing and
retrieval-augmented adaptive fusion are used to combine multimodal and
retrieved information. These systems demonstrate the value of agent
decomposition, reflection, and retrieval for emotion understanding. However,
their verification or collaboration mechanisms are designed primarily to
improve the prediction itself rather than to mechanically validate cited
evidence, localize a wrong final prediction to a responsible agent, and use that
attribution for targeted prompt repair.

\paragraph{Benchmarks for emotion capability and hallucination.}
Complementary work has focused on evaluating MLLMs rather than proposing a
new reasoning pipeline. EmoBench-M \citep{hu2025emobenchm} evaluates
multimodal emotional intelligence across foundational emotion recognition,
conversational emotion understanding, and socially complex emotion analysis.
EmotionHallucer \citep{xing2026emotionhallucer} specifically studies emotion
hallucinations through adversarial binary question answering over emotion
psychology knowledge and real-world multimodal perception. These benchmarks
provide useful model-level diagnosis of emotion understanding and hallucination,
but they evaluate model outputs rather than the internal trace of a multi-agent
reasoning process. This leaves a separate process-level problem: when an
agentic emotion system is wrong, the system still needs a way to verify the
evidence path, identify the responsible component, and route a targeted repair.


\section{Methodology}
\label{sec:method}

This section is organized into two main parts. Section~\ref{sec:agent}
presents the \emph{agentic multimodal emotion reasoning} pipeline, in which
three modality-specialist perception agents produce grounded observations,
multimodal retrieval provides relevant precedent, and an integration agent
combines both sources into an emotional draft. An inline MLLM critic then
reviews the draft and can request one revision before the final draft is
emitted. Section~\ref{sec:verification} presents the
\emph{evidence-grounded verification and prompt-space reinforcement} stage. It first
audits incorrect predictions using deterministic and entailment checks to
attribute failures, and then uses those attributed failures for prompt-space
reinforcement through clustering, patch synthesis, and held-out validation.
Figure~\ref{fig:methodology} provides the end-to-end view of the framework.
Its Boxes~5 and~6 mark the downstream verification and prompt-space
reinforcement stages, which are expanded in Figure~\ref{fig:methodology-train}.
No model weights are updated. ``MLLM'' denotes whichever backbone model
powers every agent, the critic, and patch synthesis in a given run.

\subsection{Agentic Multimodal Emotion Reasoning}
\label{sec:agent}

\begin{figure}[t]
  \centering
  \begin{tikzpicture}
    \node[anchor=south west,inner sep=0] (methodfig) at (0,0)
      {\includegraphics[width=\linewidth]{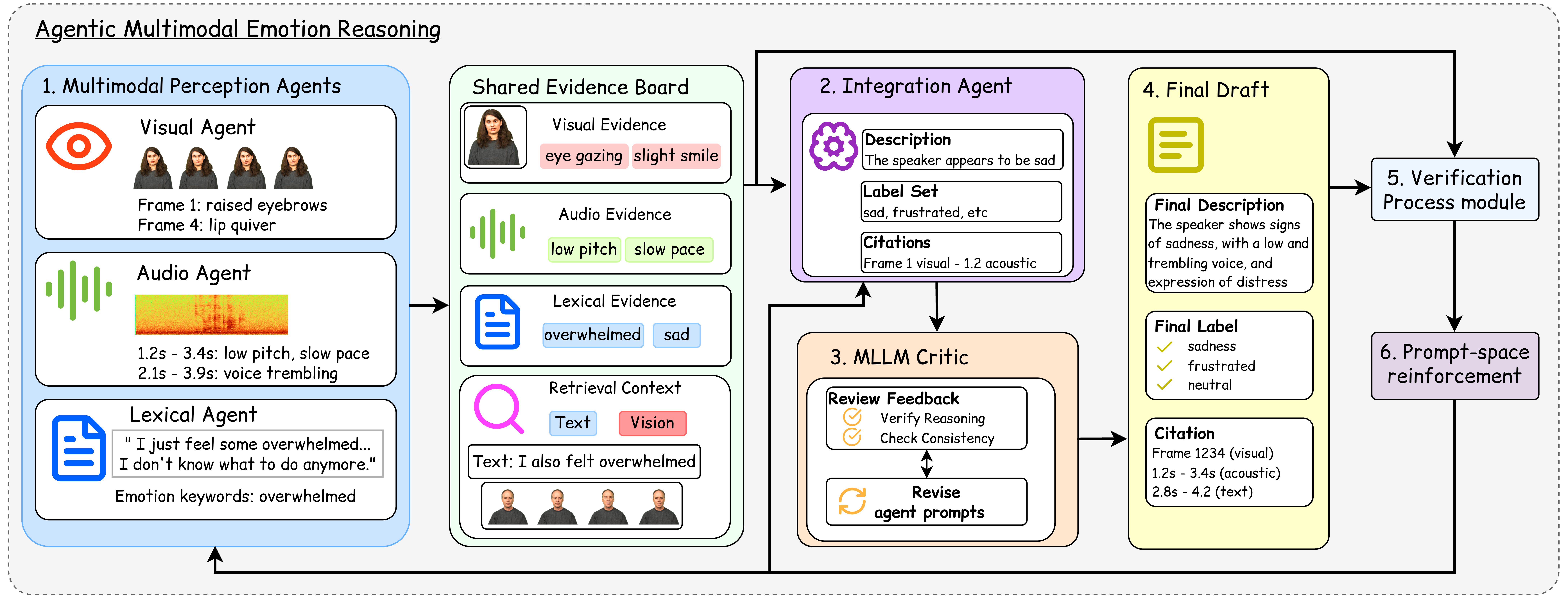}};
    \begin{scope}[x={(methodfig.south east)},y={(methodfig.north west)}]
      \node[fill=yellow!10,minimum width=1.15cm,minimum height=0.30cm,
            inner sep=0pt,font=\tiny] at (0.60,0.16) {Revise Draft};
    \end{scope}
  \end{tikzpicture}
  \caption{Agentic multimodal emotion reasoning and its connection to the downstream adaptation stages. Three modality-specialist agents post evidence-cited observations to a shared board, while multimodal retrieval provides independent text-and-vision precedent. The integration agent produces an emotional draft, and an inline MLLM critic can request one revision before the final draft is emitted. The final draft then enters the Verification Process (Box~5), whose attributed failures feed Prompt-Space Reinforcement (Box~6). Boxes~5 and~6 are expanded in Figure~\ref{fig:methodology-train}.}
  \label{fig:methodology}
\end{figure}

Each clip supplies three modality-specific inputs $X_m$, $m \in \{v, a, l\}$
(visual, acoustic, lexical), observed disjointly by the three perception
agents. The resulting evidence is combined with independently retrieved
precedent and then passed through integration and inline critique to produce
the final emotional draft, as illustrated by Boxes~1--4 in
Figure~\ref{fig:methodology}. Boxes~5 and~6 show how this draft connects to
the downstream verification and prompt-space reinforcement stages detailed in
Figure~\ref{fig:methodology-train}.

\paragraph{Multimodal perception agents.}
Three frozen agents, one that watches, one that listens, and one that reads,
each observe $X_m$ and post grounded observations to a shared evidence board,
\begin{equation}
o_m = \mathrm{Agent}_m(X_m) = \{(c_i, e_i)\}_{i=1}^{n_m}, \qquad e_i \in \mathcal{E}_m,
\label{eq:perception}
\end{equation}
pairing every claim $c_i$ with an evidence pointer $e_i$ from a
modality-specific citation space $\mathcal{E}_m$, namely a frame identifier
for $\mathcal{E}_v$, a time span for $\mathcal{E}_a$, or a transcript span for
$\mathcal{E}_l$. The \emph{visual} agent receives up to eight uniformly sampled
frames and must anchor each observation to a specific frame identifier. An
offline preprocessing stage segments the clip by Automatic Speech Recognition
(ASR). For each segment, it computes five numeric descriptors, namely pitch
mean, pitch variance, Root-Mean-Square (RMS) energy, speech rate, and the length
of the pause before the segment. The \emph{acoustic} agent reasons over these
precomputed numbers as text, citing the time span of the segment each
observation describes. The \emph{lexical} agent reads the transcript and quotes
verbatim the span it interprets. Together, their outputs form the shared
evidence board $B = \{o_v, o_a, o_l\}$. Full prompts for all three perception
agents are given in Appendix~\ref{app:prompts}
(Listings~\ref{lst:visual-prompt}--\ref{lst:lexical-prompt}).

\paragraph{Multimodal retrieval.}
Perception alone cannot name an emotion. The agents report modality-specific
signals, such as a furrowed brow, a rising pitch, or a repeated pronoun, and
turning these signals into a label requires precedent. Retrieval supplies that
precedent in the RAG tradition \citep{lewis2020rag}. An offline index over
$\sim$114k labeled samples from MER-Caption \citep{lian2025affectgpt} stores,
per sample, a text embedding of its emotion description, a visual embedding of its mean-pooled
keyframes, and metadata (label, description, transcript snippet). At inference,
the clip's own transcript and keyframe embedding query the index through two
independent nearest-neighbor searches, fused as
\begin{equation}
s(x, n_j) = \lambda \, \mathrm{sim}_{\text{text}}(x, n_j) + (1-\lambda)\, \mathrm{sim}_{\text{vis}}(x, n_j), \qquad
\mathcal{N}_k = \operatorname*{top\text{-}}k_{n_j \in \mathcal{M}} \; s(x, n_j),
\label{eq:retrieval}
\end{equation}
where $\mathcal{M}$ is the offline index and $\lambda = 0.5$ weighs the two
similarities equally. We use $k=5$; the top-$5$ neighbors $\mathcal{N}_5$ are returned as
text containing the label and description. This ensures that even a visually
driven match provides a textual precedent for the integration agent. The query
is independent of the perception agents' output, so retrieval and perception
serve as two independent evidence sources that the integration policy must
reconcile.

\paragraph{Integration agent.}
The integration agent reconciles the modality-specific observations and
retrieved precedent into an emotion. It reads the shared evidence board
$B = \{o_v, o_a, o_l\}$ alongside the retrieved neighbors $\mathcal{N}_5$,
which act as few-shot exemplars showing how comparable signal patterns have
previously mapped onto emotion labels. It runs an explicit cross-modal conflict
test and may issue one clarifying query to a single perception agent. Its
output is an emotional \emph{draft}
\begin{equation}
D = (\hat{d}, \hat{\ell}, \hat{c}, \mathcal{R}) = \mathrm{Integrator}(B, \mathcal{N}_k),
\label{eq:draft}
\end{equation}
consisting of an open-vocabulary description $\hat{d}$, a label set
$\hat{\ell}$, a confidence $\hat{c}$, and a citation list $\mathcal{R}$
resolving to specific observations or retrieved neighbors. The full prompt is
given in Appendix~\ref{app:prompts}, Listing~\ref{lst:integrator-prompt}.

\paragraph{MLLM critic.}
The critic runs inline, inside the perceive--retrieve--integrate--critic loop,
before ground truth is ever consulted. At round $t$, it reads the current draft
and the board,
\begin{equation}
v_t = \mathrm{Critic}(D_t, B) = (\mathrm{pass}_t, a_t, \iota_t),
\label{eq:critic}
\end{equation}
returning a pass or fail verdict, a suspected culprit $a_t$, and a revision
instruction $\iota_t$. On failure, the integrator redrafts under that instruction
using the same board and retrieved neighbors as before,
\begin{equation}
D_{t+1} = \mathrm{Integrator}(B, \mathcal{N}_k \mid \iota_t), \qquad t < T_{\max},
\label{eq:revise}
\end{equation}
and the critic re-checks the result, up to a fixed cap of $T_{\max} = 1$
revision. This loop only gates what the pipeline outputs. The full critic prompt
is given in Appendix~\ref{app:prompts}, Listing~\ref{lst:critic-prompt}.

\paragraph{Final draft.}
After the critic passes the draft or the revision cap is reached, the resulting
output is treated as the final draft, written simply as $D$ from here on. This
output is passed to the Verification Process shown as Box~5 in
Figure~\ref{fig:methodology}; its attributed failures can then drive the
Prompt-Space Reinforcement stage shown as Box~6. Both stages are expanded in
Figure~\ref{fig:methodology-train} and described in
Section~\ref{sec:verification}. The critic's own verdict is logged but never
feeds the later blame assignment.

\subsection{Verification Process}
\label{sec:verification}

Once the agentic reasoning stage produces the final draft $D$, it enters the
two downstream stages summarized as Boxes~5 and~6 in
Figure~\ref{fig:methodology} and expanded in
Figure~\ref{fig:methodology-train}. Box~5 is the evidence-grounded
verification process, and Box~6 uses the attributed failures for prompt-space
reinforcement. Importantly, verification is performed only on
\emph{wrong predictions}, since its purpose is to
identify a traceable failure that can later be assigned to a responsible
stage and used for prompt-space reinforcement. Correct predictions are not
audited or used for failure attribution. Formally,
\begin{equation}
\mathcal{C}(D) =
\begin{cases}
\mathrm{Det}(D) \cup \mathrm{Ent}(D), & \hat{y} \neq y^{*},\\
\varnothing, & \hat{y} = y^{*},
\end{cases}
\label{eq:gate}
\end{equation}
where $\hat{y}$ is the predicted label set $\hat{\ell}$, $y^{*}$ is the
gold label set, $\mathrm{Det}(D)$ denotes the deterministic checks, and
$\mathrm{Ent}(D)$ denotes the entailment check. The two components serve
different purposes. Deterministic checks identify mechanically verifiable
failures such as hallucinated citations or invalid references, whereas the
entailment check evaluates whether the semantic content of the draft is
actually supported by its cited evidence.

\subsubsection{Deterministic Checks}
\label{sec:deterministic-checks}

The deterministic checks are non-LLM, rule-based safeguards designed
primarily to detect mechanical hallucinations in the reasoning trace rather
than judge the semantic correctness of the predicted emotion. In particular,
they verify that citations, evidence references, retrieved examples, and
other mechanically checkable claims made by the agents correspond to
information that actually exists in the underlying sample. To keep the audit
causally relevant, clue-level checks are applied only to evidence that the
final draft actually cites; an uncited clue cannot have directly supported
the wrong prediction.

\begin{figure}[t]
  \centering
  \includegraphics[width=\linewidth]{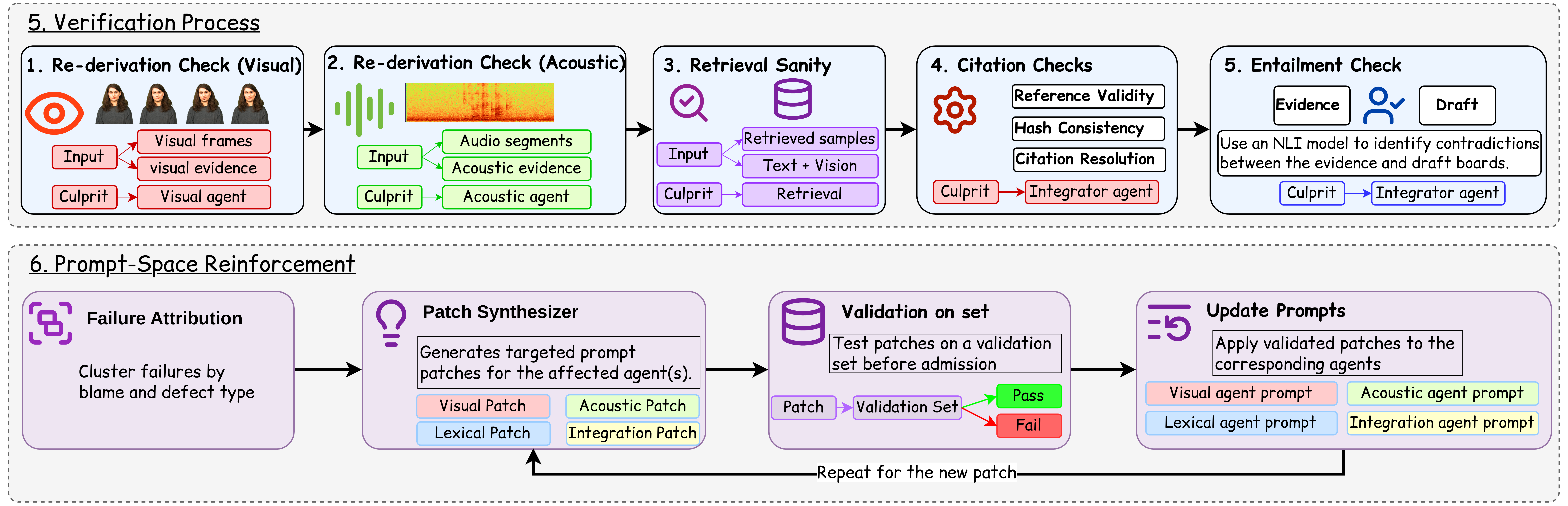}
  \caption{Detailed view of Boxes~5 and~6 from Figure~\ref{fig:methodology}. The Verification Process (Box~5) applies visual and acoustic re-derivation, retrieval sanity, citation checks, and an entailment check to the finalized draft, attributing detectable failures to a responsible stage. Prompt-Space Reinforcement (Box~6) clusters attributed failures, synthesizes targeted prompt patches, validates them on held-out data, and updates only the corresponding agent prompts when a patch is admitted.}
  \label{fig:methodology-train}
\end{figure}

Following the organization in Figure~\ref{fig:methodology-train}, the checks
cover four groups. \emph{(1) Visual re-derivation} is applied to cited
face-related visual observations and verifies that the referenced frame
exists and contains a detectable face. It therefore prevents the visual
agent from grounding a facial-expression claim in a frame where no face is
present, without attempting to determine whether the claimed expression
itself is correct. \emph{(2) Acoustic re-derivation} independently
recomputes the acoustic properties of the cited speech segment and checks
that numerical prosodic values reported by the acoustic agent, including
pitch, RMS energy, and speech rate, agree with the recomputed values within
a relative tolerance of $15\%$. \emph{(3) Retrieval sanity} checks that
retrieved neighbors satisfy a minimum similarity requirement and detects
the clear failure case in which the retrieved examples are unanimously
opposite in emotional valence to the evidence board. Finally,
\emph{(4) citation checks} verify three forms of provenance consistency.
Reference validity confirms that a cited frame, audio interval, or transcript
span actually exists in the source sample; hash consistency confirms that a
cited clue has not changed after being posted to the evidence board; and
citation resolution confirms that every identifier cited by the draft
corresponds to an actual board entry or retrieved neighbor.

Together, these checks establish that the evidence pathway used by the
draft is structurally valid and traceable. They intentionally do not decide
whether a statement such as ``the speaker appears frustrated'' is a correct
interpretation of that evidence. That content-level question is handled
separately by the entailment check.

\subsubsection{Entailment Check}
\label{sec:entailment-check}

While the deterministic checks verify the structural validity of the
evidence pathway, they cannot determine whether the semantic claims in the
draft are actually justified by that evidence. We therefore apply a separate
entailment check to the same wrong predictions. The draft description is
first decomposed into atomic claims, and each claim $c_j$ is evaluated
against the evidence $e_j$ cited in support of it. A claim passes only when
\begin{equation}
\mathrm{Entail}(c_j,e_j)
=
\mathbb{1}
\left[
    \mathrm{score}(c_j,e_j) \geq \tau_{\mathrm{ent}}
\right]
= 1,
\qquad
\tau_{\mathrm{ent}} = 0.5,
\label{eq:entail}
\end{equation}
where $\mathbb{1}[\cdot]$ is the indicator function. In contrast to the
deterministic checks, this step evaluates the \emph{content} of the claim:
the cited evidence must support the statement rather than merely exist or
resolve to a valid source.

We implemented two entailment scorers: a pretrained NLI model
(DeBERTa-v3-large \citep{he2021debertav3}) and an MLLM prompted to make a
strict grounding judgment. We evaluated both against 56 human-labeled
claim--evidence pairs, where each achieved 84\% agreement with the human
labels. We use the NLI model in the final pipeline because it provides the
same observed agreement without requiring an additional MLLM call.

The deterministic and entailment results are combined into the check set
$\mathcal{C}(D)$. A failed check provides concrete evidence for the
subsequent failure-attribution stage, which determines which component of
the agentic reasoning pipeline should receive a corrective prompt patch.
If the prediction is wrong but none of the verification checks exposes a
traceable failure, the instance is treated as irreducible and is excluded
from prompt patching.

\subsubsection{Prompt-Space Reinforcement}
\label{sec:prompt-reinforcement}

The verification process identifies recurrent, traceable failures in the
agentic reasoning pipeline. We use these failures to improve the system in
\emph{prompt space}, without updating the parameters of the underlying MLLM.
As illustrated in Figure~\ref{fig:methodology-train}, this process consists of
four steps: failure attribution, patch synthesis, validation, and prompt
update. The objective is to convert repeated verification failures into
targeted instructions for the component responsible for them, while preventing
prompt changes that degrade performance on unseen examples.

\paragraph{Failure attribution and clustering.}
Each failed verification instance is associated with the component indicated
by the corresponding verification check, such as the visual agent, acoustic
agent, retrieval module, or integration agent. Failures are then grouped by
their responsible component and defect type. This allows the system to
distinguish recurring failure patterns from isolated errors. We generate a
prompt patch only when at least $\kappa=5$ failures of the same type accumulate
for a component within a training round. Wrong predictions for which
verification finds no traceable failure are treated as irreducible and are
excluded from prompt adaptation.

\paragraph{Patch synthesis.}
For each sufficiently large failure cluster, an MLLM-based patch synthesizer
examines representative examples of the detected problem and generates a
short corrective instruction for the affected component. The patch is
targeted specifically to the failure that produced the cluster. For example,
a recurring visual grounding failure produces a patch for the visual agent,
whereas an integration-level grounding failure produces a patch for the
integration agent. The underlying model weights remain frozen; adaptation is
performed only by modifying the corresponding agent prompt. If a rule for the
same defect type already exists, the newly synthesized rule replaces it rather
than adding a duplicate instruction.

\paragraph{Patch validation and prompt update.}
A synthesized patch is treated as a candidate rather than being immediately
added to the pipeline. Testing it on the same failures from which it was
generated would provide little evidence that the rule generalizes. We
therefore insert the candidate patch into an isolated copy of the affected
agent's prompt and rerun the complete pipeline on the held-out validation set
$\mathcal{V}$. The candidate is evaluated using the same open-vocabulary
F-score used for the main evaluation and is admitted only when it does not
reduce validation performance,
\begin{equation}
\mathrm{admit}(\pi) =
\mathbb{1}
\left[
F(\mathcal{V}\mid\pi)
-
F(\mathcal{V}\mid\text{current})
\geq 0
\right].
\label{eq:admit}
\end{equation}

If the candidate passes validation, it is added to the prompt of the
corresponding component and becomes part of the policy used in subsequent
rounds. If it fails, the patch is discarded and the previous prompt is kept
unchanged. The rejected patch is also retained as feedback for the next
synthesis attempt for the same failure type, helping the synthesizer avoid
repeating an edit that has already been shown to hurt validation performance.
This produces an iterative loop in which recurring verified failures generate
candidate patches, but only validated patches are allowed to modify the
agentic reasoning pipeline.

\paragraph{Training setup.}
Prompt-space reinforcement is performed over three disjoint labeled training
rounds of approximately 331 samples each. A separate 60-sample development
split is used exclusively for validating candidate prompt patches, while a
480-sample test split is reserved for final evaluation. To study whether
targeted failure attribution is important, we compare three prompt-routing
strategies under the same patch budget. \emph{Targeted} applies each validated
patch only to the component associated with the corresponding failure cluster.
\emph{Uniform} pools the validated patches into a shared list that is provided
to all agents, removing component-specific routing. \emph{Global} also pools
the patches, but expresses them as a single free-text reflection rather than
as individual corrective rules, resembling a Reflexion-style
\citep{shinn2023reflexion} adaptation strategy.

\section{Results}
\label{sec:results}

We introduce the experimental setup, including the dataset and evaluation metrics. We then report the main results across four different MLLMs, followed by an ablation study of the framework's components and a learning-curve analysis across models.

\subsection{Evaluation Setup}
\label{sec:setup}

\paragraph{Dataset.}
We evaluate on MER2025 Track~3 (MER-DES) \citep{lian2025mer2025}, an open-vocabulary multimodal
emotion recognition benchmark. Each instance is a short video clip (around 4 seconds long) with speech, facial expression, and scene context, labeled with a free-form emotion description rather than a fixed label set. The benchmark's 1,532 manually labeled samples are split into the three training rounds, development split, and test split described in Section~\ref{sec:prompt-reinforcement}. The test split is disjoint from training and from the $114{,}395$-sample memory pool used for retrieval, so the pipeline never retrieves a test instance at inference time.

\paragraph{Evaluation Metrics.}
Labels are open-vocabulary free text, so exact string matching fails on synonyms (e.g., ``anger'' vs.\ ``furious''). We use the Emotion-Wheel (EW) metric from the MER/AffectGPT benchmark suite \citep{lian2025affectgpt}. Gold and predicted labels are mapped onto a fixed emotion-wheel vocabulary via synonym expansion, so synonymous predictions count as matches. We compute precision, recall, and F-score per instance, then average across five emotion wheels. Labels unrelated to the gold annotation still score zero, since the mapping credits semantic equivalence, not topical closeness.

\paragraph{Backbone Models.}
We test the framework on four MLLMs from four different model families,
spanning 2B to 38B parameters, so any benefit from targeted repair can be checked
across scale and architecture rather than on one backbone alone.
VideoLLaMA3-2B \citep{zhang2025videollama3} (2B), from Alibaba's DAMO
Academy, is the smallest model tested, representing a resource-constrained
deployment setting. Phi-3.5-Vision \citep{abdin2024phi3} (4.2B), from
Microsoft, targets low-latency, memory-constrained use. Qwen2.5-VL-7B
\citep{bai2025qwen25vl} (7B), from Alibaba's Qwen team, provides a mid-sized vision-language backbone,
allowing us to examine performance across model scale and architecture.
InternVL2.5-38B \citep{chen2024internvl25} (38B), from OpenGVLab, is the
largest model tested.

\subsection{Main Results}
\label{sec:rq1}

\paragraph{Motivation.}
To show the effectiveness of our targeted repair mechanism, we compare it against the static pipeline and different reflection mechanisms. This isolates the specific claim the paper depends on, that routing a fix to the agent actually responsible for a failure should outperform both doing nothing and spreading the same feedback across every agent indiscriminately.

\paragraph{Setup.}
We evaluate six conditions on the test split samples including 
a single-call baseline, a Chain-of-Thought (CoT) baseline, the
static multi-agent pipeline with no repair (Boxes~1--4 in Figure~\ref{fig:methodology}), and the pipeline augmented with
three reflection modes, \emph{global} (one undifferentiated revision
broadcast to all agents), \emph{uniform} (the same patch content applied to
every agent regardless of blame), and \emph{targeted} (patches routed only
to the blamed agent). All three repair modes are trained for three rounds
on the same training data before evaluation. We report the official
open-vocabulary label metric (F-score, precision, recall). Conditions 3--6 all run the same five agent prompts given in Appendix~\ref{app:prompts}, differing only in what repair content (if any) is injected into them; conditions 4--6 additionally receive the reflection or targeted-repair text produced by that condition's training rounds. Conditions 1 and 2 instead use the single-call Direct and CoT prompts (Appendix~\ref{app:prompts}, Listings~\ref{lst:direct-prompt} and \ref{lst:cot-prompt}), which bypass the agent decomposition entirely.

\paragraph{Results.}

\begin{table*}[t]
\centering
\footnotesize
\caption{Official label metric (\%) across all six conditions, repeated for four MLLMs ordered by parameter count, on the full 480-sample test split. F: F-score, P: Precision, R: Recall. Bold marks the best value per column.}
\label{tab:rq1-main}
\begin{tabular*}{\textwidth}{@{\extracolsep{\fill}}lcccccccccccc@{}}
\toprule
& \multicolumn{3}{c}{VideoLLaMA3-2B} & \multicolumn{3}{c}{Phi-3.5-Vision} & \multicolumn{3}{c}{Qwen2.5-VL-7B} & \multicolumn{3}{c}{InternVL2.5-38B} \\
\cmidrule(lr){2-4}\cmidrule(lr){5-7}\cmidrule(lr){8-10}\cmidrule(lr){11-13}
Condition & F & P & R & F & P & R & F & P & R & F & P & R \\
\midrule
1: Direct            & 35.6 & \textbf{45.1} & 29.4 & 41.1 & 52.0 & 33.9 & 46.3 & 58.0 & 38.6 & 53.3 & 63.0 & 46.2 \\
2: CoT                & 31.0 & 41.5 & 24.7 & 38.0 & 49.5 & 30.8 & 44.1 & 55.0 & 36.7 & 54.5 & 63.5 & 47.7 \\
3: Static pipeline    & 39.0 & 41.9 & 36.5 & 47.6 & 50.1 & 45.4 & 52.3 & 54.6 & 50.3 & 59.1 & 61.0 & 57.1 \\ \midrule
4: + Global refl.     & 38.2 & 41.6 & 35.4 & 47.1 & 49.7 & 44.8 & 52.2 & 54.8 & 49.7 & 58.7 & 60.8 & 56.8 \\
5: + Uniform refl.    & \textbf{39.6} & 42.4 & \textbf{37.0} & 48.1 & 50.8 & \textbf{45.7} & 53.4 & 56.0 & \textbf{51.0} & 60.0 & 62.1 & 58.1 \\
6: + Targeted repair  & \textbf{39.6} & 43.0 & 36.7 & \textbf{49.0} & \textbf{53.0} & 45.6 & \textbf{54.5} & \textbf{58.9} & 50.9 & \textbf{61.2} & \textbf{64.5} & \textbf{58.3} \\
\bottomrule
\end{tabular*}
\end{table*}

Table~\ref{tab:rq1-main} reports all six conditions across four MLLMs.
F-score rises monotonically with model scale, and the gap between the
smallest and largest model is large: targeted repair reaches only $39.6$ F
on VideoLLaMA3-2B against $61.2$ F on InternVL2.5-38B.
Global reflection scores below the static pipeline on all four models, so broadcasting undifferentiated feedback can hurt more than it helps. Uniform reflection and targeted repair both beat the static pipeline everywhere. Targeted repair gets the best F-score on three of the four models and ties uniform reflection on the fourth (VideoLLaMA3-2B, $39.6$ F). The repair effect also scales with the backbone: the targeted-versus-static delta is $2.2$ F for Qwen2.5-VL-7B and $2.1$ F for
InternVL2.5-38B, but only $0.6$ F for VideoLLaMA3-2B, since blame
attribution and patch synthesis are both done by the model being corrected, and a 2B model is likely too weak to critique itself. Precision and recall diverge: targeted repair wins precision on three of the four models, but on VideoLLaMA3-2B the direct baseline has the highest precision instead ($45.1$ vs.\ $43.0$). CoT prompting only helps the largest model: it
underperforms the direct baseline on VideoLLaMA3-2B, Phi-3.5-Vision, and
Qwen2.5-VL-7B, and only edges past it on InternVL2.5-38B. Smaller models lack the capacity to reason through multiple steps without losing
the correct label, and only InternVL2.5-38B is large enough for
reasoning to help rather than hurt.


\subsection{Ablation Study}

\paragraph{Motivation.}
Targeted repair combines several removable components, shown across
Figures~\ref{fig:methodology} and~\ref{fig:methodology-train}. To show how much each one actually matters, rather than assuming
the full pipeline is necessary as built, we ablate them one at a time and
measure the resulting drop in test performance.
 
\paragraph{Setup.}
We run the ablation on InternVL2.5-38B, the best-performing model, using the
targeted-repair condition. Each row removes exactly one component, the
critic, memory retrieval, deterministic checks, the entailment check,
clustering, or the validation gate's re-evaluation step, and evaluates the
resulting system on the same 480-sample test split. We report F-score,
precision, recall, and $\Delta F$ relative to the full pipeline.

\begin{table}[t]
\centering
\caption{Component ablation of targeted repair (InternVL2.5-38B), full
480-sample test split. Each row removes one component, $\Delta F$ relative
to the full pipeline.}
\label{tab:rq3-ablation}
\begin{tabular}{lcccc}
\toprule
Ablation & F-score (\%) & $\Delta F$ (\%) & Precision (\%) & Recall (\%) \\
\midrule
Full targeted repair        & $\mathbf{61.2}$ & --      & \textbf{64.5} & 58.3 \\
\midrule
Critic                           & $60.7$ & $-0.5$ & 61.5 & \textbf{59.9} \\
Memory retrieval (RAG)           & $59.1$ & $-2.1$ & 62.0 & 56.5 \\
Deterministic checks             & $60.3$ & $-0.9$ & 63.0 & 57.8 \\
Entailment check                 & $60.0$ & $-1.2$ & 62.2 & 58.0 \\
Clustering                       & $59.4$ & $-1.8$ & 61.0 & 57.9 \\
Re-evaluation (validation gate)  & $58.2$ & $-3.0$ & 60.0 & 56.5 \\
\bottomrule
\end{tabular}
\end{table}

\paragraph{Results.}
Every ablation lowers F-score (Table~\ref{tab:rq3-ablation}), with the magnitude of the drop depending on the removed component. Removing the critic causes the smallest decrease ($-0.5$ F). The drop appears entirely in precision ($-3.0$), while recall actually rises ($+1.6$). This is consistent with the critic filtering a narrow slice of false positives rather than performing independent verification. Deterministic checks cause a larger decrease ($-0.9$ F). In addition to filtering at inference time, they shape which failures accumulate enough evidence to be correctly blamed and clustered during training. The entailment check causes a $-1.2$ F drop. It is the only direct detector of semantic corruption in the verification layer. Without it, blame attribution for content-level errors degrades, resulting in less targeted patches. Memory retrieval and clustering cause larger drops ($-2.1$ and $-1.8$ F, respectively). Removing retrieval lowers both precision and recall, whereas removing clustering mainly lowers precision. Memory retrieval provides the integrator with relevant precedent, while clustering provides enough evidence from similar defects to clear the admission floor. Thus, neither component simply filters an existing draft. Removing the re-evaluation pass causes the largest decrease ($-3.0$ F). It also produces the largest precision drop ($-4.5$) and ties memory retrieval for the largest recall drop ($-1.8$). This result is consistent with its role as the pipeline's main safeguard against allowing a drifting or ineffective patch to reach later rounds without confirmation.

\subsection{Training Curve}
\label{sec:rq3}

\paragraph{Motivation.}
Labeled data for the repair mechanism is costly to collect. We want to
know how much of it targeted repair actually needs, and how its
performance changes as more data becomes available.
 
\paragraph{Setup.}
We test targeted repair at 20 training-data fractions, from 5\% to 100\% in steps of 5\%. Each fraction is a nested subset of the next larger one. We run the full repair procedure at each fraction on three models, Qwen2.5-VL-7B, Phi-3.5-Vision, and InternVL2.5-38B, and report F-score, precision, and recall on the same held-out test split for each one.
 
\paragraph{Results.}

\begin{figure}[t]
  \centering
  \includegraphics[width=\linewidth]{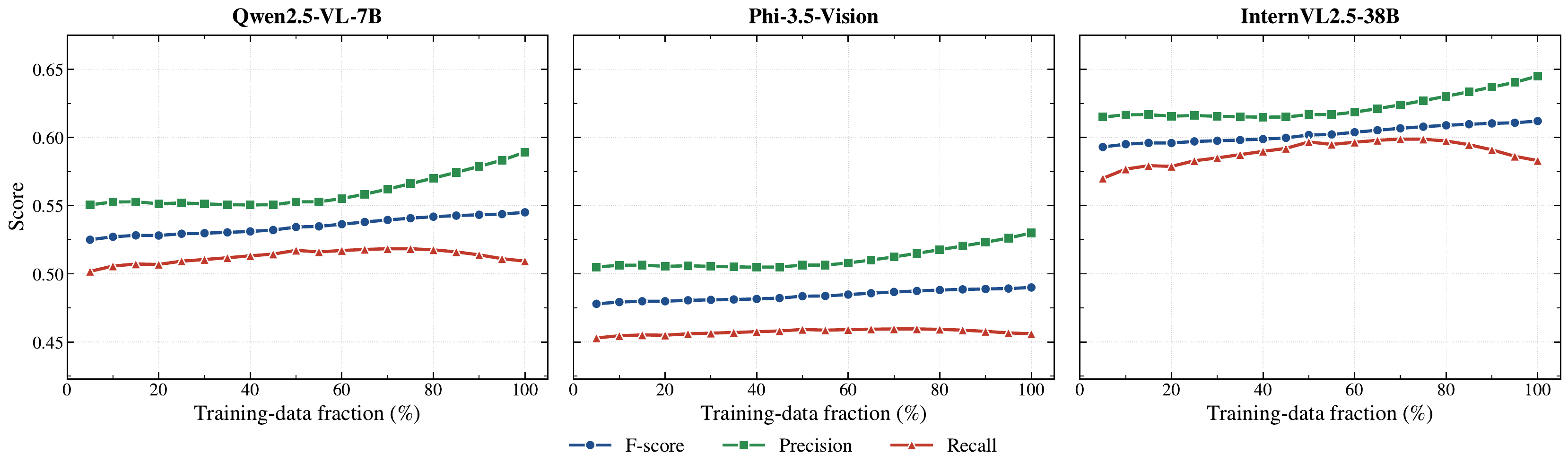}
  \caption{F-score, precision, and recall against training-data fraction,
    for three models.}
  \label{fig:curve}
\end{figure}

Figure~\ref{fig:curve} shows the same overall pattern for all three
models. F-score climbs steadily as training data grows, with no
meaningful decline at any fraction. This is a good sign. It means the repair mechanism is not overfitting to the training pool, and more labeled data would likely help further. The gain from 5\% to 100\% of the data is modest in absolute terms, $+2.0$ F for Qwen2.5-VL-7B, $+1.9$ F for InternVL2.5-38B, and $+1.2$ F for Phi-3.5-Vision, so a small labeled set already captures most of the benefit, but performance keeps climbing rather than plateauing or falling back.
 
Precision and recall move differently. Precision stays close to flat for
roughly the first half of the range, then climbs steadily from about 50\%
to 100\% of the data in all three models. Recall behaves the other way.
It rises early, peaks around 70 to 75\% of the data, then gives a little
back by 100\%. This rollover is sharpest for InternVL2.5-38B, where
recall visibly declines over the last quarter of the range. Read
together, early data mostly helps recall, later data mostly sharpens
precision, and F-score keeps improving throughout because the precision
gain outweighs the small recall loss.


\section{Limitations}
\label{sec:limitations}

Two limitations are worth noting. First, the verification layer audits
only what a draft actually cites (Section~\ref{sec:deterministic-checks}):
a wrong prediction whose citations pass every deterministic and entailment
check is labeled irreducible (Section~\ref{sec:entailment-check}) and cannot
be attributed to any agent or repaired, so the repair loop's reach is capped
by how much of the error rate is attributable in the first place, not by how
well it acts on the errors it does catch. Second, the pipeline issues
multiple sequential model calls per clip (three perceivers, retrieval, the
integrator, and up to one critic-triggered revision), compared with the
single-call Direct and CoT baselines in Table~\ref{tab:rq1-main}. We do not
report latency, throughput, or per-clip inference cost for any condition, so
the accuracy gains reported in Section~\ref{sec:rq1} come with an
unquantified overhead that matters for the real-time deployment settings
this line of work targets.

\section{Conclusion and Future Work}
\label{sec:conclusion}

We presented a three-part agentic framework for multimodal emotion
reasoning that traces a wrong prediction back to the agent responsible
for it, rather than relying on another model's judgment of plausibility:
modality-specialist perception agents post evidence-cited observations,
an integration agent drafts a description refined by an inline MLLM
critic, and a separate verification layer audits the finalized draft and
attributes failures using deterministic and entailment checks. A
verifier-guided repair loop turns these attributions into targeted,
held-out-validated prompt patches with no weight updates anywhere in the
loop. Across four MLLMs from 2B to 38B parameters on MER2025, targeted
repair outperformed a frozen static pipeline on every backbone and matched
or outperformed two undifferentiated reflection baselines, strictly
improving on both for three of the four models, with the ablation study showing
that each pipeline component contributes a measurable share of the gain.
Future work could turn the same per-agent attributions into supervision
for fine-tuning the responsible agent, and the per-claim deterministic
and entailment scores into a process-level reinforcement-learning
reward, extending the current training-free loop to drive weight updates
directly.


\bibliographystyle{unsrtnat}
\bibliography{references}

\appendix
\section{Prompts}
\label{app:prompts}

This appendix gives the exact system prompts used in the paper: the base
(round-0, pre-repair) prompts for the three modality-specialist perceivers
(Listings~\ref{lst:visual-prompt}--\ref{lst:lexical-prompt}), the
integration agent (Listing~\ref{lst:integrator-prompt}), and the critic
(Listing~\ref{lst:critic-prompt}), all used identically across conditions
3--6 (\S\ref{sec:rq1}) before any round-specific repair content is added;
and the two single-call baseline prompts used for condition~1, Direct
(Listing~\ref{lst:direct-prompt}), and condition~2, Chain-of-Thought
(Listing~\ref{lst:cot-prompt}). Each \texttt{\{...\}} placeholder is filled
in per instance at inference time (segment features, transcript text,
frame ids, board and neighbor content, and so on); the surrounding JSON
schema is otherwise fixed. Listing~\ref{lst:patch-example} then shows a
concrete instance of the verifier-guided repair described in
\S\ref{sec:prompt-reinforcement}: the constraint text that targeted repair
appended to the visual and integrator prompts above by round 3, after
credit assignment had attributed a recurring defect to each.

\begin{lstlisting}[caption={Visual perceiver prompt (round 0, base).}, label={lst:visual-prompt}]
You are the visual perceiver in a multimodal emotion recognition pipeline. You are given {num_frames} video thumbnails sampled at roughly 1 frame per second, in temporal order. The images are attached in order and correspond to frame ids: {frame_ids}.

Extract 1 to 8 visual clues relevant to the depicted person's/people's emotional state: facial expressions, body language, gestures, posture changes, and relevant scene/context details.

For each clue, cite the "frame_id" (one of: {frame_ids}) of the specific thumbnail that grounds it. Only use frame ids from that exact list.

Return ONLY a JSON object of this exact form, nothing else, no markdown code fences, no commentary:
{{"clues": [{{"content_text": "<one sentence describing the visual clue and its emotional relevance>", "frame_id": "<one of {frame_ids}>", "confidence": <float between 0 and 1>}}, ...]}}
\end{lstlisting}

\begin{lstlisting}[caption={Acoustic perceiver prompt (round 0, base).}, label={lst:acoustic-prompt}]
You are the acoustic perceiver in a multimodal emotion recognition pipeline. You do not hear the audio directly -- you are given precomputed acoustic features per speech segment, extracted deterministically from the waveform and ASR transcript.

Segments:
{segments_block}

For EACH segment listed above, write exactly one acoustic clue interpreting what those numbers suggest about the speaker's emotional state (e.g. high f0 variance and fast speech rate may indicate excitement or agitation; a long pause_before may indicate hesitation).

Your content_text MUST start with this exact fixed prefix for that segment, with the three numbers copied verbatim (same digits, do not round differently or recompute), followed by a colon and your interpretation:
"f0_mean=<exact f0_mean value> Hz, rms_mean=<exact rms_mean value>, speech_rate=<exact speech_rate value> words/sec: <your interpretation>"
All three numbers (f0_mean, rms_mean, speech_rate) MUST appear -- do not drop any of them even if your interpretation only discusses one or two.

You MUST cite time_span as [start, end] copied exactly from that segment.

Return ONLY a JSON object of this exact form, nothing else, no markdown code fences, no commentary:
{{"clues": [{{"content_text": "f0_mean=<value> Hz, rms_mean=<value>, speech_rate=<value> words/sec: <interpretation>", "time_span": [<start>, <end>], "confidence": <float between 0 and 1>}}, ...]}}

There must be exactly {num_segments} clues, one per segment, in the same order as listed above.
\end{lstlisting}

\begin{lstlisting}[caption={Lexical perceiver prompt (round 0, base).}, label={lst:lexical-prompt}]
You are the lexical perceiver in a multimodal emotion recognition pipeline. You are given the ASR transcript of a short video clip's spoken dialogue (it may contain transcription errors; work with what is given).

Transcript:
"{transcript}"

Extract 3 to 8 lexical clues relevant to the speaker's emotional state: semantic content, sentiment-bearing words or phrases, and discourse cues (hedges, intensifiers, repetition, rhetorical questions, etc.).

For each clue, quote a "transcript_span": a VERBATIM substring copied exactly from the transcript above (same characters, same case, same punctuation) that grounds the clue. Do not paraphrase or translate the span -- copy it exactly as it appears in the transcript.

Return ONLY a JSON object of this exact form, nothing else, no markdown code fences, no commentary:
{{"clues": [{{"content_text": "<one sentence describing the clue and its emotional relevance>", "transcript_span": "<verbatim substring from the transcript>", "confidence": <float between 0 and 1>}}, ...]}}
\end{lstlisting}

\begin{lstlisting}[caption={Integration agent prompt (round 0, base).}, label={lst:integrator-prompt}]
You are the integrator in a multimodal emotion recognition pipeline. You read clues already posted to a shared board by three perceivers (visual, acoustic, lexical) plus similar examples retrieved from a memory index, and produce one open-vocabulary emotion description with labels.

Board clues (cite by clue_id):
{board_block}

Retrieved similar examples (cite by neighbor_id; these are examples from other clips, not this clip -- use them only as context for typical emotion-label vocabulary and patterns, never as direct evidence about this clip):
{neighbors_block}

First, decide whether the clues conflict across modalities -- for example, positive/agreeable words spoken in an angry tone, or a smiling face paired with hostile speech content, can indicate sarcasm or irony rather than the literal sentiment of any single modality.

Completeness requirement: if an acoustic clue is posted on the board, your description must either cite it and account for what it says, or explicitly note in conflict_explanation why it was set aside (e.g. contradicted by stronger visual/lexical evidence). Never make a claim about tone, pitch, or vocal delivery unless you cite the acoustic clue that supports it -- acoustic evidence is the modality most likely to disambiguate a case that looks resolved from visual+lexical alone (e.g. sarcasm, forced calm, suppressed anger), so it must be actively reconciled, not skimmed.

Then write a Draft: a short natural-language description of the person's emotional state grounded in the clues, a list of open-vocabulary emotion labels, and a citations list containing the clue_id and/or neighbor_id of every board clue or retrieved example you actually used. Every citation must be one of the ids listed above -- do not invent ids.

Only if a specific perceiver's clue is genuinely ambiguous or insufficient to resolve a conflict, you may request ONE follow-up from exactly one perceiver (visual, acoustic, or lexical) by setting "query". Most clips do not need this -- leave query as null unless there is a real, specific ambiguity you cannot resolve from the clues already given.

Return ONLY a JSON object of this exact form, nothing else, no markdown code fences, no commentary:
{{
  "conflict_detected": <true or false>,
  "conflict_explanation": "<one sentence, or null if conflict_detected is false>",
  "draft": {{
    "description_text": "<natural-language description of the emotional state>",
    "labels": ["<open-vocab emotion word>", ...],
    "citations": ["<clue_id or neighbor_id actually used>", ...],
    "confidence": <float between 0 and 1>
  }},
  "query": {{"agent": "visual|acoustic|lexical", "instruction": "<specific follow-up instruction>"}} or null
}}
\end{lstlisting}

\begin{lstlisting}[caption={Critic prompt (round 0, base). The critic is never itself a target of prompt-space reinforcement in the current design (\S\ref{sec:prompt-reinforcement}), so this prompt does not change across rounds.}, label={lst:critic-prompt}]
You are the critic in a multimodal emotion recognition pipeline. You review a Draft produced by the integrator against the clues it was built from, and decide whether it should pass.

Board clues:
{board_block}

Draft under review:
- description_text: {description_text}
- labels: {labels}
- citations: {citations}
- confidence: {confidence}
{deterministic_checks_block}

Check whether the draft's description and labels are actually supported by the cited clues (not just plausible in general), whether every citation refers to a real clue, and whether the confidence is reasonable given the evidence. If deterministic check results are provided above, treat any failing check as a strong signal the draft should fail.

If the draft fails, name exactly one specific agent responsible (visual, acoustic, lexical, or integrator) and give one concrete, actionable revision instruction naming the specific defect (e.g. "the acoustic clue about high pitch is not reflected in the labels" or "citation X does not support the claim of anger").

Return ONLY a JSON object of this exact form, nothing else, no markdown code fences, no commentary:
{{"passed": <true or false>, "suspected_culprit": "<visual|acoustic|lexical|integrator|retrieval_gap|irreducible or null if passed>", "revision_instruction": "<specific instruction naming the agent and defect, or null if passed>"}}
\end{lstlisting}

\begin{lstlisting}[caption={Direct single-call baseline prompt, condition 1 (\S\ref{sec:rq1}). No agent decomposition, no board, no retrieval.}, label={lst:direct-prompt}]
You are analyzing a short video clip for emotion recognition. You are given the clip's spoken-dialogue transcript and one representative thumbnail image.

Transcript:
"{transcript}"

Directly predict the open-vocabulary emotion label(s) expressed by the person in this clip, and a short description of the emotional evidence.

Return ONLY a JSON object of this exact form, nothing else, no markdown code fences, no commentary:
{{"description_text": "<short description of the emotional evidence>", "labels": ["<open-vocab emotion word>", ...], "confidence": <float between 0 and 1>}}
\end{lstlisting}

\begin{lstlisting}[caption={Chain-of-Thought (CoT) single-call baseline prompt, condition 2 (\S\ref{sec:rq1}). Same inputs as the Direct baseline (Listing~\ref{lst:direct-prompt}); only the elicited reasoning step differs.}, label={lst:cot-prompt}]
You are analyzing a short video clip for emotion recognition. You are given the clip's spoken-dialogue transcript and one representative thumbnail image.

Transcript:
"{transcript}"

Think step by step before answering: consider what the words, tone, and visible expression each suggest, and whether they agree or conflict. Then predict the open-vocabulary emotion label(s) expressed by the person in this clip, and a short description of the emotional evidence.

Return ONLY a JSON object of this exact form, nothing else, no markdown code fences, no commentary. Write your step-by-step reasoning in "reasoning_text" first, then the final answer in "description_text" and "labels":
{{"reasoning_text": "<step-by-step reasoning about the lexical, visual, and vocal evidence, considering agreement or conflict between them>", "description_text": "<short description of the emotional evidence>", "labels": ["<open-vocab emotion word>", ...], "confidence": <float between 0 and 1>}}
\end{lstlisting}

\begin{lstlisting}[caption={Example verifier-guided patches (\S\ref{sec:prompt-reinforcement}) admitted for the targeted-repair condition by round 3. Each block was appended verbatim to the corresponding round-0 prompt above (Listings~\ref{lst:visual-prompt} and \ref{lst:integrator-prompt}) once its defect cluster passed the admission floor (Eq.~\ref{eq:admit}).}, label={lst:patch-example}]
[Visual, appended to Listing 1 -- defect type: visual_unverified_face_claim]
Patches from verifier-attributed failure analysis (apply as hard constraints):
- [visual_unverified_face_claim_r3] Ensure that all cited visual claims are supported by actual face detection in the corresponding frames, and that the described facial expressions align with the cited emotion labels, while preserving plausible label coverage through multimodal evidence integration.

[Integrator, appended to Listing 4 -- consolidated guidance for this agent, derived from verifier-attributed failures]
Ensure that each produced description accurately reflects the multimodal evidence by cross-referencing all evidence types with the predicted labels, avoiding unsupported claims that do not align with the evidence. Additionally, ensure that the predicted labels align with the multimodal evidence and avoid contradictions with the retrieved neighbors' labels, specifically by reconciling any discrepancies between the lexical, visual, and acoustic data and the retrieved neighbors' labels.
\end{lstlisting}


\newpage
\section*{NeurIPS Paper Checklist}

\begin{enumerate}

\item {\bf Claims}
    \item[] Question: Do the main claims made in the abstract and introduction accurately reflect the paper's contributions and scope?
    \item[] Answer: \answerYes{}
    \item[] Justification: The abstract and Section~\ref{sec:intro} state the three-part agentic framework (perception, verification, verifier-guided repair), the MER2025 evaluation across four MLLMs, and the F-score gains, all of which are substantiated in Sections~3--4 and the ablation study.

\item {\bf Limitations}
    \item[] Question: Does the paper discuss the limitations of the work performed by the authors?
    \item[] Answer: \answerYes{}
    \item[] Justification: A dedicated Limitations section (\S\ref{sec:limitations}) discusses the irreducible class of errors that verification cannot attribute or repair, and the unquantified inference-time cost of the multi-call pipeline relative to the single-call baselines.

\item {\bf Theory assumptions and proofs}
    \item[] Question: For each theoretical result, does the paper provide the full set of assumptions and a complete (and correct) proof?
    \item[] Answer: \answerNA{}
    \item[] Justification: The paper is an empirical systems contribution; it defines its pipeline and metrics formally (Eqs.~1--12) but does not present theorems or formal proofs.

    \item {\bf Experimental result reproducibility}
    \item[] Question: Does the paper fully disclose all the information needed to reproduce the main experimental results of the paper to the extent that it affects the main claims and/or conclusions of the paper (regardless of whether the code and data are provided or not)?
    \item[] Answer: \answerYes{}
    \item[] Justification: Section~3 gives the full pipeline as explicit equations (Eqs.~1--12), Table~1 lists all six deterministic checks, \S3.4 specifies the training rounds, split sizes, and routing regimes, \S4.1 specifies the dataset, metric, and backbone models, and Appendix~\ref{app:prompts} gives the exact system prompts (Listings~1--8) used by every agent.

\item {\bf Open access to data and code}
    \item[] Question: Does the paper provide open access to the data and code, with sufficient instructions to faithfully reproduce the main experimental results, as described in supplemental material?
    \item[] Answer: \answerYes{}
    \item[] Justification: An anonymized code repository is linked in the abstract (\url{https://anonymous.4open.science/r/Multi_Agent_Emotion_Reasoning-CF5A/README.md}); the evaluation dataset (MER2025 Track~3) is a public benchmark.

\item {\bf Experimental setting/details}
    \item[] Question: Does the paper specify all the training and test details (e.g., data splits, hyperparameters, how they were chosen, type of optimizer) necessary to understand the results?
    \item[] Answer: \answerYes{}
    \item[] Justification: \S3.4 specifies the three training rounds ($\sim$331 samples each), the 60-sample development split, and the 480-sample test split; \S4.1 specifies the retrieval index size, the four backbone MLLMs, and fixed thresholds ($\tau=0.15$, $\tau_{\mathrm{ent}}=0.5$, $\kappa=5$, $\epsilon_F=0$) used throughout. No gradient-based optimizer is used, since the method is training-free (prompt-space only).

\item {\bf Experiment statistical significance}
    \item[] Question: Does the paper report error bars suitably and correctly defined or other appropriate information about the statistical significance of the experiments?
    \item[] Answer: \answerNo{}
    \item[] Justification: Tables~2--3 and Figure~2 report single-run point estimates (F-score, precision, recall) on the fixed 480-sample test split without error bars, confidence intervals, or multi-seed variance.

\item {\bf Experiments compute resources}
    \item[] Question: For each experiment, does the paper provide sufficient information on the computer resources (type of compute workers, memory, time of execution) needed to reproduce the experiments?
    \item[] Answer: \answerNo{}
    \item[] Justification: The paper does not report GPU/hardware type, memory, or wall-clock time for the perception, verification, or repair stages.
    
\item {\bf Code of ethics}
    \item[] Question: Does the research conducted in the paper conform, in every respect, with the NeurIPS Code of Ethics \url{https://neurips.cc/public/EthicsGuidelines}?
    \item[] Answer: \answerYes{}
    \item[] Justification: The work uses an existing public benchmark (MER2025 Track~3) and frozen, publicly released MLLMs; it introduces no new human-subject data collection beyond a small internal check of the entailment scorer against 56 human-labeled claims (\S3.2).

\item {\bf Broader impacts}
    \item[] Question: Does the paper discuss both potential positive societal impacts and negative societal impacts of the work performed?
    \item[] Answer: \answerNo{}
    \item[] Justification: The paper does not include a societal-impact discussion. Emotion recognition carries dual-use risk (e.g., non-consensual affect inference or surveillance), which is not addressed in the current draft.
    
\item {\bf Safeguards}
    \item[] Question: Does the paper describe safeguards that have been put in place for responsible release of data or models that have a high risk for misuse (e.g., pre-trained language models, image generators, or scraped datasets)?
    \item[] Answer: \answerNA{}
    \item[] Justification: The paper releases prompt patches and pipeline code, not a new pretrained model, dataset, or generator with elevated misuse risk.

\item {\bf Licenses for existing assets}
    \item[] Question: Are the creators or original owners of assets (e.g., code, data, models), used in the paper, properly credited and are the license and terms of use explicitly mentioned and properly respected?
    \item[] Answer: \answerNo{}
    \item[] Justification: The MER2025 dataset and the four backbone MLLMs (VideoLLaMA3, phi-3.5-vision, Qwen2.5-VL, InternVL2.5) are cited by their original papers, but the paper does not state their license names or terms of use explicitly.

\item {\bf New assets}
    \item[] Question: Are new assets introduced in the paper well documented and is the documentation provided alongside the assets?
    \item[] Answer: \answerNA{}
    \item[] Justification: The paper does not release a new dataset or trained model; the pipeline code referenced in the abstract will be released upon de-anonymization.

\item {\bf Crowdsourcing and research with human subjects}
    \item[] Question: For crowdsourcing experiments and research with human subjects, does the paper include the full text of instructions given to participants and screenshots, if applicable, as well as details about compensation (if any)? 
    \item[] Answer: \answerNo{}
    \item[] Justification: \S3.2 reports that the entailment scorer was checked against 56 human-labeled claims, but the paper does not describe annotator instructions or compensation for that labeling.

\item {\bf Institutional review board (IRB) approvals or equivalent for research with human subjects}
    \item[] Question: Does the paper describe potential risks incurred by study participants, whether such risks were disclosed to the subjects, and whether Institutional Review Board (IRB) approvals (or an equivalent approval/review based on the requirements of your country or institution) were obtained?
    \item[] Answer: \answerNA{}
    \item[] Justification: The paper does not conduct new human-subject research; the only human involvement is a small internal annotation check of the entailment scorer (\S3.2) on an existing public benchmark's clips, not a study with recruited participants.

\item {\bf Declaration of LLM usage}
    \item[] Question: Does the paper describe the usage of LLMs if it is an important, original, or non-standard component of the core methods in this research? Note that if the LLM is used only for writing, editing, or formatting purposes and does \emph{not} impact the core methodology, scientific rigor, or originality of the research, declaration is not required.
    \item[] Answer: \answerYes{}
    \item[] Justification: MLLMs are the core methodological component, not a writing aid: Section~3 and Appendix~\ref{app:prompts} describe in full how frozen MLLMs act as the perception, integration, critic, and patch-synthesis agents underlying the proposed framework.

\end{enumerate}

\end{document}